# Learning to Be a Transformer to Pinpoint Anomalies

**ALEX COSTANZINO, PIERLUIGI ZAMA RAMIREZ, GIUSEPPE LISANTI, AND LUIGI DI STEFANO, (Member, IEEE)**

Department of Computer Science and Engineering (DISI), Computer Vision Laboratory (CVLAB), University of Bologna, 40126 Bologna, Italy

Corresponding author: Alex Costanzino (alex.costanzino@unibo.it)

This work was supported in part by "FSE+ 2021–2027 ai sensi dell'art. 24, comma 3, lett. a), della Legge 240/2010 e s.m.i. e del D.G.R. 693/2023 (RIF. PA: 2023-20090/RER) under Grant CUP: J19J23000730002" and in part by SACMI Imola SC.

**ABSTRACT** To efficiently deploy strong, often pre-trained feature extractors, recent Industrial Anomaly Detection and Segmentation (IADS) methods process low-resolution images, e.g., $224 \times 224$ pixels, obtained by downsampling the original input images. However, while numerous industrial applications demand the identification of both large and small defects, downsampling the input image to a low resolution may hinder a method's ability to pinpoint tiny anomalies. We propose a novel Teacher–Student paradigm to leverage strong pre-trained features while processing high-resolution input images very efficiently. The core idea concerns training two shallow MLPs (the Students) by nominal images so as to mimic the mappings between the patch embeddings induced by the self-attention layers of a frozen vision Transformer (the Teacher). Indeed, learning these mappings sets forth a challenging pretext task that small-capacity models are unlikely to accomplish on out-of-distribution data such as anomalous images. Our method can spot anomalies from high-resolution images and runs way faster than competitors, achieving state-of-the-art performance on MVTec AD and the best segmentation results on VisA. We also propose novel evaluation metrics to capture robustness to defect size, i.e., the ability to preserve good localisation from large anomalies to tiny ones. Evaluating our method also by these metrics reveals its neatly superior performance.

## I. INTRODUCTION

Industrial anomaly detection and segmentation (IADS) aims to identify anomalous samples and localise their defects. This task is challenging in industrial applications, where anomalies are diverse and unpredictable, and there may be a limited number of nominal samples. Indeed, in IADS the training process is typically unsupervised, with the training set consisting solely of nominal samples. Modern approaches for IADS [1], [2], [3], [4], [5], [6], [7], [8] create a model of the nominal images during training. Then, at inference time, each test image is compared to this nominal model, and any discrepancy is interpreted as an anomaly.

To meet the timing requirements of industrial applications while leveraging strong, often pre-trained, feature extractors, current IADS methods resort to processing low-resolution images, such as, e.g., $224 \times 224$, obtained by downsampling the original input images. However, this path is detrimental to sensitivity since small anomalies may be lost due to strong downsampling (fig. 1).

Conversely, we propose a novel IADS paradigm that relies on rich, pre-trained feature extractors, enabling the efficient processing of high-resolution images. As such, our methodology is less prone to missing tiny anomalies while keeping up with strict timing constraints concerning both training and inference. We deploy a frozen Transformer encoder alongside a novel Teacher–Student paradigm whereby two shallow MLPs (the Students), shared across patch embeddings, learn to mimic the contextualization – and decontextualization – transformations realised by the layers of the backbone (the Teacher) while observing nominal images. The core intuition behind this idea is that, upon optimisation, the small-capacity

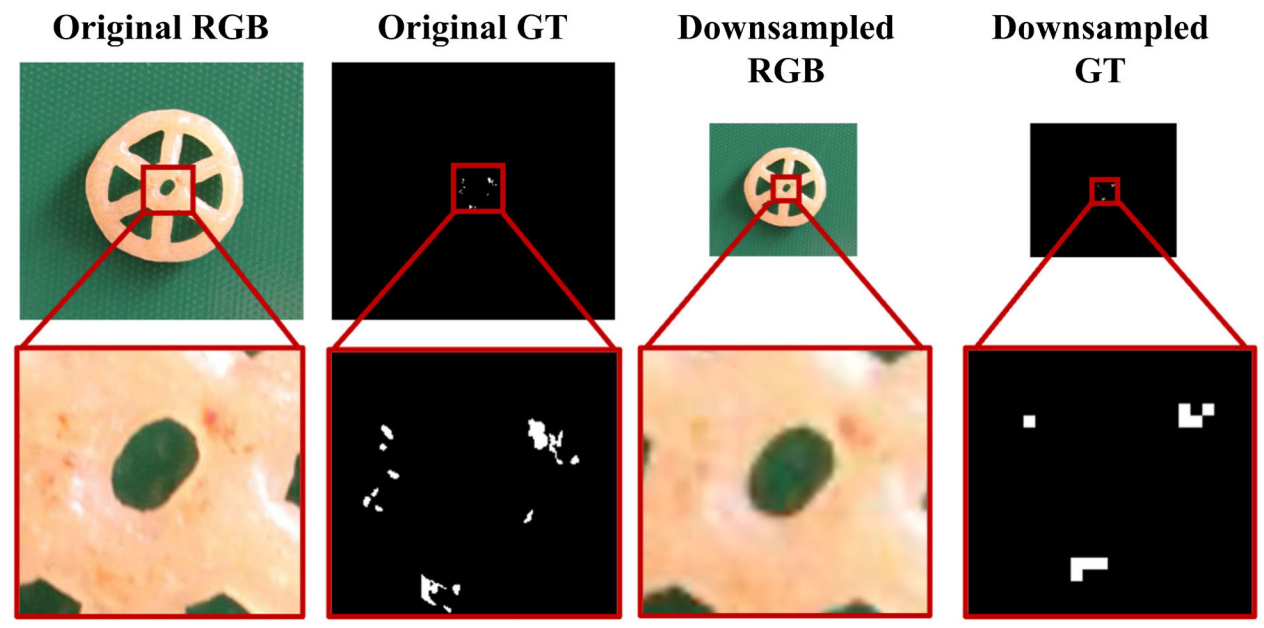

**FIGURE 1.** Effects of downsampling on VisA [9]. Tiny defects are no longer visible in both the input image and the ground-truth.

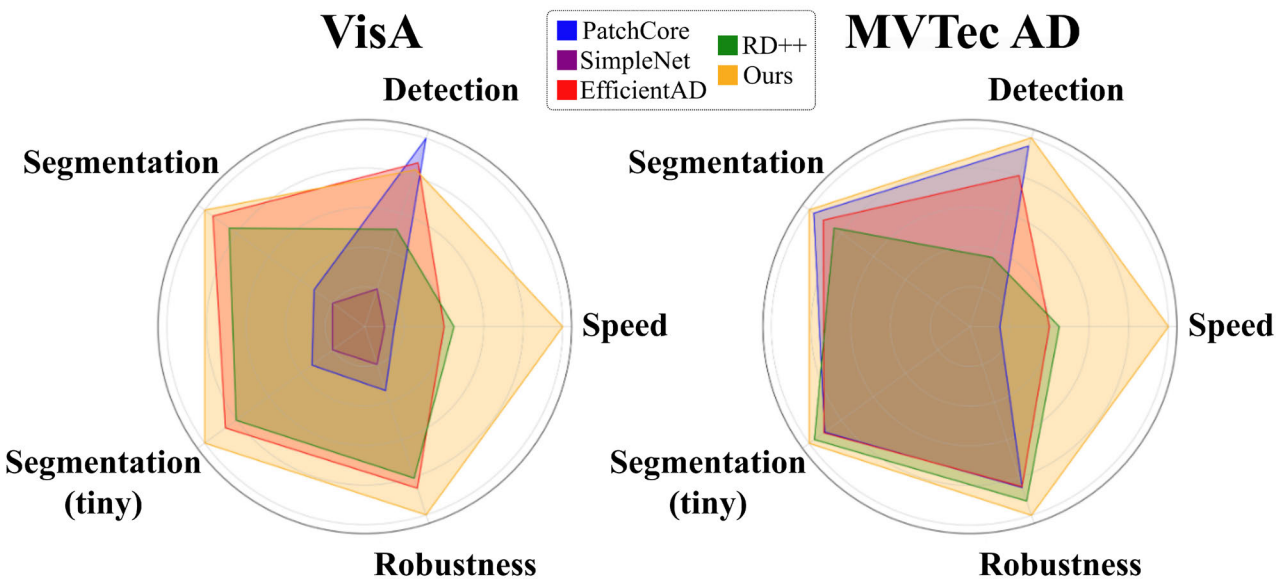

**FIGURE 2.** Comparison between IADS methods. The metrics reported in the charts are describedin Sec. IV. Values are normalized for better readability.

Student networks will be able to hallucinate contextual information from more local content – and vice versa – on nominal samples only, while faltering to do so when presented with anomalous ones. Thus, at inference time, the actual features computed by the Teacher on each image patch are compared to those predicted by the Students, with the discrepancies between the former and the latter highlighting anomalies.

Our Teacher–Student paradigm is general and can be applied effectively to any frozen Transformer backbone. However, the best performance is obtained by learning the Students on the features extracted by DINO-v2 [10], which has been trained with images of varying sizes, including high-resolution images. In such a configuration, our method compares favourably to previous proposals on the reference IADS benchmarks [9], [11] while running notably faster, for instance, so as to scrutinise $1036 \times 1036$ images in $\sim$2 ms on a NVIDIA GeForce RTX 4090. Key to the speed of our method is its reliance on shallow Students shared across patches, which enables extremely fast processing due to the patch embeddings being handled as independent samples during both inference and training. Focusing on the latter, we reckon that, as the samples to train the Students are the patch embeddings, each nominal image provides multiple training samples, e.g., 5476 for a $1036 \times 1036$ image, which vastly extends the training set size compared to the number of training images and hints that our method holds the potential to achieve excellent performance also in challenging few-shot regimes.

Finally, we point out that it is common practice in the literature to downsample also the ground-truth masks provided by IADS benchmarks to match the resized, low-resolution input images. Accordingly, defect masks get smaller and tiny anomalies may disappear from the ground-truth (fig. 1). Hence, the practice described above might not accurately reflect the ability of current methods to pinpoint defects of all sizes. Differently, in our experimental evaluation we rely on the original, high-resolution ground-truth masks and, to highlight the benefits that come with processing high-resolution inputs, we propose novel evaluation metrics to measure segmentation performance relative to the size of anomalies and to capture robustness with respect to defect sizes, i.e., the ability to preserve segmentation performance from large anomalies to smaller ones. Evaluating our method with this novel protocol reveals its ability to detect even tinier defects better than competitors as well as its superior robustness to the size of anomalies.

Our contributions can be summarised as follows.

- We propose a novel IADS method that relies on training two shallow MLPs to mimic the transformations realised by the self-attention layers of a frozen vision Transformer;
- Our approach yields state-of-the-art detection and segmentation performance on MVTec AD [11] and state-of-the-art segmentation results on VisA [9] while running at a remarkably faster speed than all competitors (fig. 2);
- We introduce novel evaluation metrics to assess how effectively and robustly IADS methods can handle anomalies of different sizes, showing that our approach neatly outperforms competitors (fig. 2);
- We propose a challenging few-shot setting built upon the VisA dataset and find that our proposal achieves state-of-the-art segmentation performance.

## II. RELATED WORK

### A. ANOMALY DETECTION SOLUTIONS

Solutions for IADS can be categorised based on the approach followed to model nominal samples. Normalizing Flows [2], [3], [12], [13], [14] based methods construct complex distributions by transforming a probability density via a series of invertible mappings. In particular, these methods extract features of normal images from a pre-trained model and transform the feature distribution into a Gaussian distribution during the training phase. At test time, after passing the extracted features through the Normalizing Flow, the features of abnormal images will deviate from the Gaussian distribution of the training, suggesting an anomaly. Lately, several solutions [1], [15], [16] that employ Memory Banks have been introduced. This category of solutions exploits well-known feature extractors trained on a large plethora of data [10], [17], [18] to model nominal samples. More in detail, during training, the feature extractor is kept frozen and used to compute features for nominal samples, which are then stored in a memory bank. At test time, the features extracted from an input image are compared to those in the bank to identify anomalies. Despite their remarkable performance, these approaches suffer from slow inference speed, since each feature vector extracted from the input image needs to be compared against all the nominal feature vectors stored in the memory bank. Methods close to our solution which follows a Teacher–Student strategy [5], [6], [7], [8], [19], [20], [21] have also been proposed. In this family of solutions, the training phase involves a Teacher model that extracts features from nominal samples and distils this knowledge to the Student model, which learns to mimic the Teacher's feature extraction process. During the testing phase, differences between the features generated by the Teacher model and those produced by the Student model reveal the presence of anomalies. Recently, a multimodal-only approach, CFM [22], investigated the idea of mapping features from one modality to the other on nominal samples and then detecting anomalies by pinpointing inconsistencies between observed and mapped features. This solution leverages MLPs to learn a mapping between features coming from two different modalities, RGB images and point clouds. Unlike CFM [22], our novel solution does not require two modalities. Peculiarly, our technique runs significantly faster than all existing methods while achieving superior performance, especially when employing high-resolution images, thanks to our novel design of the Teacher-Student paradigm that involves lightweight MLPs as student networks. We can appreciate the advantages of our method over competitors in Table 2 and Figure 2.

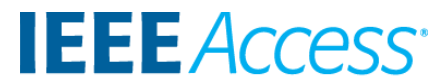

**FIGURE 3.** **Method overview.Given an RGB Image $I$, a frozen transformer backbone $\mathcal{T}$ is leveraged to extract two sequences of patch-aligned features $F_j$, $F_k$, from different layers, one from a shallower layer $j$, and one from a deeper layer $k$.Then, a pair of MLPs, $\mathcal{S}_F$, $\mathcal{S}_B$, predict the extracted features from one layer to the other, processing the features at each patch independently.Extracted and predicted features are compared to obtain two layer-specific anomaly maps, $\Delta_j$, $\Delta_k$, that are then combined to obtain the final anomaly map $\Delta$.**

### B. ANOMALY DETECTION BENCHMARKS

During the last few years, several IADS datasets have been released. The introduction of MVTec AD [11] kicked off the development of AD&S approaches for industrial applications. This dataset contains several industrial inspection scenarios, each comprising both train and test sets. Each train set contains only nominal images, while the test sets also contain anomalous samples. Such a scenario represents realistic real-world applications where types and possible locations of defects are unknown during the development of AD&S algorithms. Later, the work was extended with the MVTec 3D-AD [23] dataset, which follows the same structure of MVTec AD, but also provides the pixel-aligned point clouds of the samples to address the AD&S in a multimodal fashion. Shortly afterwards, the Eyecandies [24] dataset was released, which mimics the structure of MVTec 3D-AD by introducing a multimodal synthetic dataset containing images, point clouds and normals for each sample. To provide a more challenging scenario, the VisA dataset [9] has been introduced, in which high-resolution images of complex scenes that can also contain multiple instances of the same object have been released. In the end, more task-specific datasets such as MAD [25] and MVTec LOCO [26] have been released. In particular, MAD [25] introduced a multi-pose dataset with images from different viewpoints covering a wide range of poses to tackle a pose-agnostic AD&S. MVTec LOCO [26] not only contains structural anomalies, such as dents or holes, but also logical anomalies, which violations of logical constraints can be, for instance, a wrong ordering or a wrong combination of normal objects.

## III. METHOD

As outlined in fig. 3, our method follows a Teacher-Student paradigm in which the Teacher, $\mathcal{T}$, is a frozen Transformer encoder (e.g., DiNO-v2 [10]), while the two Students, referred to as Forward and Backward Students, $\mathcal{S}_F$ and $\mathcal{S}_B$, are realised as shallow MLPs.

### A. OVERVIEW

The Students are trained on nominal samples and learn to mimic the transformations between the patch embeddings occurring within the layers of the Transformer. In particular, the Forward Student $\mathcal{S}_F$ learns to predict the patch embeddings computed by a layer of the Transformer ($k$ in fig. 3), given the corresponding embeddings computed by a previous layer ($j$ in fig. 3). Conversely, the Backward Student $\mathcal{S}_B$ learns to predict the features calculated by the Transformer at layer $j$ given the corresponding ones at layer $k$. The Student networks are shared across patch embeddings, i.e., both take as input the features associated with the patch $(i)$ at a layer $f_j^{(i)}$, $f_k^{(i)}$ and predict the corresponding features at the other layer $\hat{f}_k^{(i)}, \hat{f}_j^{(i)}$. At inference time, for all patch embeddings of the given test sample, the features predicted by the Students are compared to those extracted by the Teacher, with the discrepancies between the former and the latter providing the signal to highlight anomalies. As shown in fig. 3, the differences between the outputs from $\mathcal{S}_F$, $\mathcal{S}_B$ and the patch embeddings from layers $k$,$j$ of $\mathcal{T}$ yield two anomaly maps, $\Delta_j$ and $\Delta_k$, that are fused to obtain the final one.

### B. RATIONALE

The intuition behind our approach relies on the observation that as patch embeddings travel from shallower to deeper layers of a Transformer encoder, they become increasingly contextualised, i.e., deeper representations capture more global information that helps to single out a patch based on the specific context provided by the input image. Our Students are trained to contextualise and decontextualise patch embeddings according to the function, which we assume to be invertible, executed by the Transformer between a pair of chosen layers. In particular, contextualization gathers and integrates local information to form more global and coherent patch representations, whilst decontextualization realises the opposite process, i.e., highlights the inherent local features from more global representations. By learning contextualization and decontextualization on nominal samples, our Students will understand how local features, such as edges and textures, fit into larger structures, such as shapes and objects, in nominal entities. We conjecture that feature contextualization and decontextualization are complex functions that do not admit a trivial solution, e.g., the identity function. Therefore, small-capacity networks trained only on nominal samples are unlikely to learn general functions that can yield correct predictions on out-of-distribution data, i.e., features extracted from anomalous patches. Hence, when presented with anomalous samples, the learned global vs. local mappings break down because the local and global features predicted by the Students do not align with those computed by the Teacher, revealing inconsistencies that pinpoint anomalies.

### C. TEACHER

As a first step, we provide as input to the Teacher $\mathcal{T}$ an image $I$ with dimensions $H \times W \times C$, where $H$, $W$, and $C$ correspond to the height, width, and number of channels. In our framework, we employ a Transformer-based backbone that provides a set of features, one for each input patch processed by the backbone after each layer. Each feature, $f^{(i)} \in \mathbb{R}^D$, has dimension $D$ according to the inner representation employed by the backbone, while the number of features is $N = HW/P^2$, where the patch size is $P \times P$ pixels. During the forward pass, we extract two sets of features, $F_j = \left\{f_j^{(i)}, i = 1 \cdots N\right\}$ and $F_k = \left\{f_k^{(i)}, i = 1 \cdots N\right\}$, from two different layers of the backbone, i.e., layers $j$ and $k$, with $j < k$.

We highlight that, as far as the representation of small defects is concerned, a Transformer backbone can effectively handle high-resolution inputs because, although it processes images by dividing them into patches, which results in smaller spatial size, the input information is not compressed, on the contrary, each patch is expanded to a higher dimensionality related to the internal representation of the Transformer (e.g., RGB patches of 14 × 14 × 3 pixels are mapped into 768-dimensional embeddings). Therefore, as high-resolution information is retained, we can also detect small defects.

### D. STUDENTS

The two sets of features extracted by the Teacher are processed by a pair of lightweight networks, $\mathcal{S}_F$ and $\mathcal{S}_B$, representing the Students in our architecture. $\mathcal{S}_F$ maps a feature vector from a less contextualized layer $j$ to a more contextualized layer $k$, while $\mathcal{S}_B$ does the opposite. Each network predicts the features of one layer from the corresponding ones extracted from the other, processing each patch location independently. Thus, given a patch location $(i)$ and the corresponding features $f_j^{(i)}$ and $f_k^{(i)}$, the features predicted by the Students can be expressed as:

$$\hat{f}_k^{(i)} = \mathcal{S}_F(f_j^{(i)}) \quad \hat{f}_j^{(i)} = \mathcal{S}_B(f_k^{(i)}) \tag{1}$$

where $\mathcal{S}_F$ and $\mathcal{S}_B$ are parametrised as MLPs, shared across all patches. By processing all patches, we obtain the two sets of predicted features: $\hat{F}_j = \left\{\hat{f}_j^{(i)}, i = 1 \cdots N\right\}$ and $\hat{F}_k = \left\{\hat{f}_k^{(i)}, i = 1 \cdots N\right\}$.

As stated in Sec. I, employing Student networks that process each patch independently with shallow MLPs enables fast inference. Moreover, as each patch is in an independent training sample, this approach effectively increases the training set size relative to the number of training images. Consequently, our method can be trained on a few images while achieving excellent performance (see Sec. V).

### E. TRAINING

During training, the weights of $\mathcal{S}_F$ and $\mathcal{S}_B$ are optimised only on nominal samples of a specific class from a dataset. For both networks, we employ the cosine distance between the features extracted from the backbone at the considered layers and the transferred ones as a loss function. Thus, the per-patch losses are:

$$\begin{aligned} \mathcal{L}_j^{(i)}\left(f_j^{(i)}, \hat{f}_j^{(i)}\right) &= 1 - \frac{f_j^{(i)} \cdot \hat{f}_j^{(i)}}{\|f_j^{(i)}\| \|\hat{f}_j^{(i)}\|} \\ \mathcal{L}_k^{(i)}\left(f_k^{(i)}, \hat{f}_k^{(i)}\right) &= 1 - \frac{f_k^{(i)} \cdot \hat{f}_k^{(i)}}{\|f_k^{(i)}\| \|\hat{f}_k^{(i)}\|} \end{aligned} \tag{2}$$

The complete training algorithm is devised in algorithm 1.

### F. INFERENCE

At inference time, the image under analysis is processed by the Transformer backbone and the features extracted from the

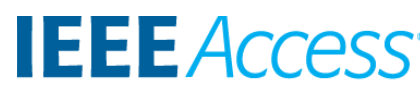

**Algorithm 1** Training Algorithm

**Require:** Pre-trained Teacher network $\mathcal{T}$;
**Require:** Sequence of training images $\{I_n\}$.
1: Initialize Student networks $\mathcal{S}_F$ and $\mathcal{S}_B$;
2: **for** $I \in \{I_n\}$ **do**
3: Extract patch features $f_j^{(i)}$, $f_k^{(i)} \leftarrow \mathcal{T}(I)$ with the Teacher network $\mathcal{T}$ at layers $j$ and $k$;
4: Predict patch features $\hat{f}_k^{(i)} \leftarrow \mathcal{S}_F(f_j^{(i)})$ from layer $j$ to $k$ with the Forward Student $\mathcal{S}_F$ and $\hat{f}_j^{(i)} \leftarrow \mathcal{S}_B(f_k^{(i)})$ from $k$ to $j$ with the Backward Student $\mathcal{S}_B$;
5: Back-propagate the per-patch reconstruction losses $\mathcal{L}_j^{(i)}$ and $\mathcal{L}_k^{(i)}$ to optimize the Student networks.
6: **end for**

two layers, $F_j$ and $F_k$ are provided as input to the Students networks to obtain the corresponding predicted features, $\hat{F}_j$ and $\hat{F}_k$. The Euclidean distance is then employed to compute the patch-wise differences between extracted and predicted features $\Delta_j^{(i)}$, $\Delta_k^{(i)}$:

$$\Delta_j^{(i)} = \|f_j^{(i)} - \hat{f}_j^{(i)}\|_2, \quad \Delta_k^{(i)} = \|f_k^{(i)} - \hat{f}_k^{(i)}\|_2, \tag{3}$$

with $\Delta_j = \left\{\Delta_j^{(i)}, i = 1 \cdots N\right\}$, $\Delta_k = \left\{\Delta_k^{(i)}, i = 1 \cdots N\right\}$. Typically, we can identify anomalies from both $\Delta_j$ and $\Delta_k$, i.e., from both prediction directions. However, in case of failure of one Student network, the bidirectional mapping creates a fail-safe mechanism since it is unlikely for an anomaly to pass through contextualization and decontextualization without detection. Thus, we fuse the predicted anomaly maps $\Delta_j$ and $\Delta_k$ by multiplying the difference values corresponding to the same patch, such that: $\Delta^{(i)} = \Delta_j^{(i)} \cdot \Delta_k^{(i)}$, with $\Delta = \left\{\Delta^{(i)}, i = 1 \cdots N\right\}$. Finally, the set of fused differences, $\Delta$, is reshaped as a $\sqrt{N} \times \sqrt{N}$ anomaly map according to the positions of the patches within the input image. This map is then up-sampled to $H \times W$, i.e., the spatial size of the input image, by bilinear interpolation and successively smoothed according to common practice [1], [7], [22], [27]. The global anomaly score required to perform sample-level anomaly detection is computed as the mean value of the top $M$ values of the final anomaly map. The complete inference algorithm is devised in algorithm 2.

**Algorithm 2** Inference Algorithm

**Require:** Pre-trained Teacher network $\mathcal{T}$;
**Require:** Trained Student networks $\mathcal{S}_F$ and $\mathcal{S}_B$;
**Require:** Sequence of test images $\{I_n\}$.
1: **for** $I \in \{I_n\}$ **do**
2: Extract patch features $f_j^{(i)}$, $f_k^{(i)} \leftarrow \mathcal{T}(I)$ with the Teacher network $\mathcal{T}$ at layers $j$ and $k$;
3: Predict patch features $\hat{f}_k^{(i)} \leftarrow \mathcal{S}_F(f_j^{(i)})$ from layer $j$ to $k$ with the Forward Student $\mathcal{S}_F$ and $\hat{f}_j^{(i)} \leftarrow \mathcal{S}_B(f_k^{(i)})$ from $k$ to $j$ with the Backward Student $\mathcal{S}_B$;
4: Compute patch-wise difference between extracted and predicted features $\Delta_j^{(i)}$, $\Delta_k^{(i)}$ at layers $j$ and $k$;
5: Aggregate the patch-wise differences $\Delta^{(i)} = \Delta_j^{(i)} \cdot \Delta_k^{(i)}$;
6: Upsample $\Delta$ at the original input dimension and smooth it with a Gaussian blur;
7: Compute the global anomaly score as the mean of the top $M$ values of $\Delta$.
8: **end for**

## IV. EXPERIMENTAL SETTINGS

To assess our proposal, we rely on two popular IADS datasets: VisA [9] and MVTec AD [11]. The VisA [9] dataset provides images of varying resolution, with the height spanning from 1284 to 1562 pixels and anomalies as tiny as 1 pixel and up to 478781 pixels. The dataset contains 10821 images of 12 objects across 3 domains, with challenging scenarios including complex structures in objects, multiple instances, and pose variations. Between the provided images, 9621 are nominals while 1200 contain defects. The MVTec AD dataset includes 5354 images, with heights spanning from 700 to 1024 pixels and anomalies ranging from 24 pixels to 517163 pixels. The images pertain to 15 objects exhibiting 73 different types of anomalies for 1888 anomalous samples. Both datasets provide pixel-accurate ground-truths for each anomalous sample.

As highlighted in fig. 4, VisA features a significantly wider range of anomaly sizes and includes tiny defects. As a result, downsampling the ground-truths to 224 × 224 pixels, i.e., the most commonly employed inference and evaluation size in the present literature, yields a reduction in the number of defects of 21.42% and 0.37% for VisA and MVTec AD, respectively. These observations render VisA a particularly challenging scenario for assessing the robustness of IADS methods with respect to defect size.

### A. METRICS

#### 1) STANDARD METRICS

We utilise the metrics employed in MVTec AD [11] and VisA [9]. These two datasets assess image-level anomaly detection performance by the Area Under the Receiver Operator Curve (I-AUROC) computed on the global anomaly score. As for defect segmentation performance, the Area Under the Per-Region Overlap (AUPRO) on the anomaly map is computed, with the integration threshold set to 0.3. Recently, [8], [22] have proposed to compute the AUPRO considering a tighter threshold, i.e., 0.05. We will consider both metrics and denote AUPROs with integration thresholds 0.3 and 0.05 as AUPRO@30%, and AUPRO@5%, respectively.

#### 2) PERFORMANCE ACROSS DEFECT SIZES

To highlight the capability of IADS methods to segment defects with varying sizes, we introduce a variation of the AUPRO metric. In particular, for each object in a dataset, we compute the anomaly size distribution and partition it into cumulative quartiles, denoted as $Q_1, Q_2, Q_3, Q_4$. These cumulative quartiles are associated with sets that contain only anomalies with a size smaller than or equal to the considered quartile. Hence, the set associated with Q4 consists of all anomalies, while Q1 includes only the smallest ones. Then, we calculate the AUPRO@30% and AUPRO@5% on each set, with the segmentation metrics associated with Q4 being the segmentation metrics adopted in the standard benchmarks.

#### 3) ROBUSTNESS

We also introduce a novel metric, $\rho$, to assess the robustness of a method with respect to the size of the defects in a dataset. In particular, $\rho$ captures a method's ability to segment tiny and larger defects accurately. Accordingly, we define robustness as:

$$\begin{aligned} \rho &= w \cdot (1 - s)\,, \\ w &= \frac{1}{4} \cdot \sum_{i=1}^{4} \text{AUPRO}(Q_i), \\ s &= \frac{|\text{AUPRO}(Q_4) - \text{AUPRO}(Q_1)|}{\max\big(\text{AUPRO}(Q_1), \text{AUPRO}(Q_4)\big)} \end{aligned} \tag{4}$$

Here, for the sake of compactness, we denote as AUPRO either AUPRO@5% or AUPRO@30%, such that considering the former or the latter will yield $\rho$@5% or $\rho$@30%, respectively. In the definition of $\rho$, the AUPRO is evaluated for the smallest defects only, i.e., AUPRO($Q_1$), and for all defects, i.e., AUPRO($Q_4$). With this measure, if a method can correctly segment larger defects but struggles with small ones, its sensitivity to defect size, $s$, is high, and its robustness, $\rho$, is low. Conversely, a robust method should be able to accurately segment defects regardless of their sizes, which in our metric would be captured by the difference between AUPRO($Q_4$) and AUPRO($Q_1$) turning out low, yielding low sensitivity and high robustness. Yet, to avoid deeming as robust a method that performs poorly on both small and large defects, such that AUPRO($Q_4$) and AUPRO($Q_1$) are both similarly low, in the definition of $\rho$ we introduce the average AUPRO across all quartiles, $w$, as weighing factor of the term $(1 - s)$. The proposed robustness metric, $\rho$, is bounded by 1 since both $s$ and $w$ are smaller than 1.

### B. EVALUATION PROTOCOL & IMPLEMENTATION DETAILS

#### 1) EVALUATION PROTOCOL

We evaluate our proposal alongside with several state-of-the-art IADS methods, such as PatchCore [1], SimpleNet [27], RD++ [7] and EfficientAD [8] (in particular, EfficientAD-M provides better performance).

As described in [8], the results reported in SimpleNet [27] are obtained by repeatedly evaluating the metrics on all test images during training to select the best checkpoint. Analysing the official implementation, we noticed how this protocol has also been followed by RD++ [7]. However, in real-world settings, the test data is not available at training time. Thus, to avoid overestimating the actual performance of the models, we disable the above checkpoint selection mechanism, train all methods for a fixed number of epochs and evaluate the model obtained at the last checkpoint. For [7], [8], and [27], we train for the number of epochs specified in the official implementations.

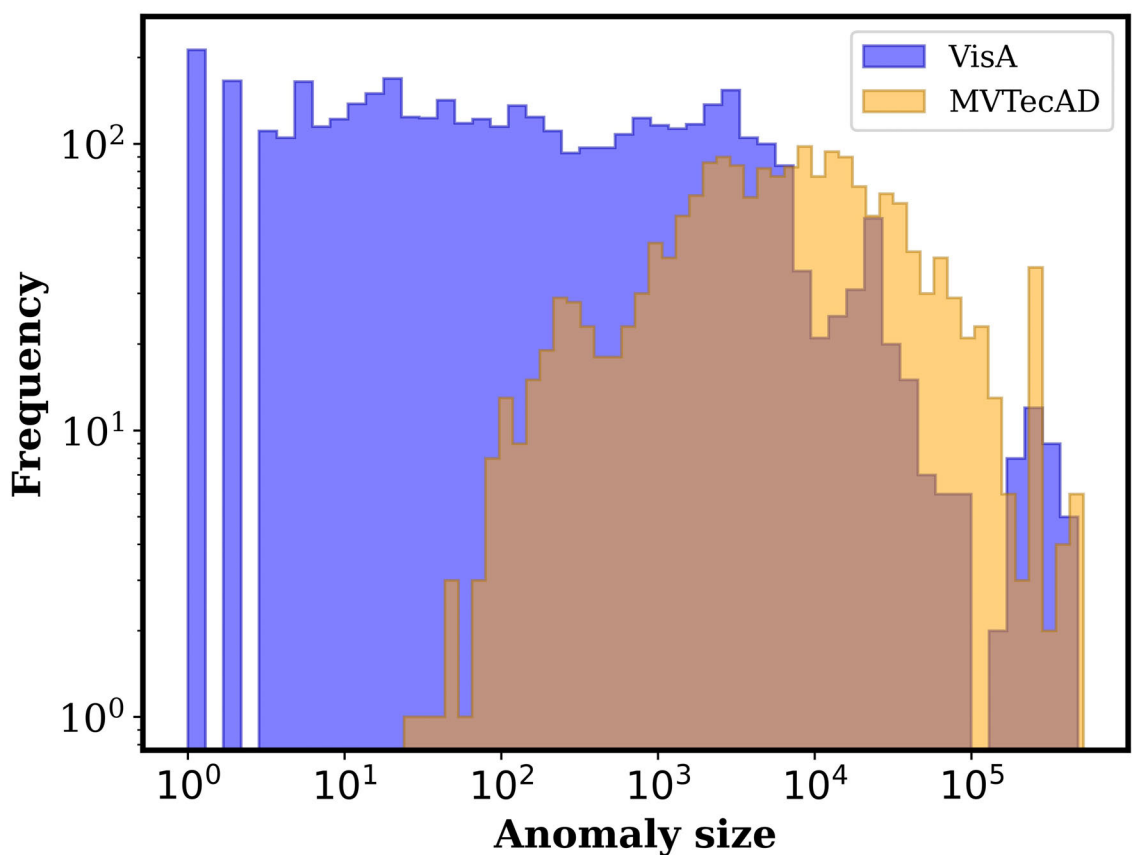

**FIGURE 4. Anomaly size distribution.**

PatchCore [1] employs a centre crop of the input images since in MVTec AD, most defects lie within this cropped area. However, in a real-world scenario, anomalies can occur outside of this area. Thus, we disable this strategy as it implies knowledge about the location of anomalies.

As anticipated in Sec. I, we compute all metrics based on the original ground-truths provided with the datasets, which have the same resolution as the original input images. Hence, we do not downsample the ground-truths to the input image size processed by a method, but we bilinearly upsample the anomaly map to the same resolution as the ground-truth in order to calculate all metrics.

Some methods, including ours, must add padding to the image to adapt it to the input size of the employed backbone. However, we remove these extra pixels from the final anomaly maps as, otherwise, they usually decrease the False Positive Rate (and thus artificially ameliorate the segmentation metrics) because they tend to yield very low anomaly scores. Finally, we calculate the AUPRO considering all the samples in the test set, both nominal and anomalous.[1]

#### 2) IMPLEMENTATION DETAILS

As our default Teacher network, we employ DINO-v2 ViT-B/14 [10] pre-trained on a large, curated, and diverse dataset of 142 million images, comprising ImageNet-22k [28], [29]. Thus, our $\mathcal{T}$ network processes $1036 \times 1036 \times 3$ RGB images and outputs $74 \times 74 \times 768$ feature maps. Both $\mathcal{S}_F$ and $\mathcal{S}_B$ consist of three linear layers, each but the last one followed by GeLU activations. The GeLU activation function – modelled as $\text{GeLU}(x) = x \cdot \Phi(x)$, where $\Phi(x)$ is the Cumulative Distribution Function for the Gaussian Distribution – has been chosen since it is the same activation function employed inside DINO-v2 layers and the goal of $\mathcal{S}_F$ and $\mathcal{S}_B$ is to mimic the transformation inside DINO-v2. The number of units per layer is 768 for both $\mathcal{S}_F$ and $\mathcal{S}_B$, to match the dimensionality of the input $F_j$, $F_k$ and output $\hat{F}_j$, $\hat{F}_k$ feature embeddings. The processing of the Students networks is shared across patches as the features at each patch are processed independently. The two networks are trained jointly for 50 epochs using Adam [30] with a learning rate of 0.001. As default, we select the layers $j = 8$ and $k = 12$ to realise the forward and backward predictions of the patch embeddings. We employed $M = 0.001 \cdot H \cdot W$ to attain the number of pixels used to calculate the global anomaly score. We report in fig. 5 an exploded view of the internal architecture of the Students networks. We conducted all the experiments on a single NVIDIA GeForce RTX 4090.

[1] We noticed that the official code from [1] calculates the AUPRO only on anomalous test samples, obtaining higher scores since the false positive rate is inherently lower with this protocol.

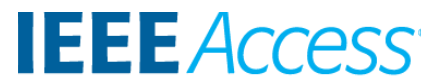

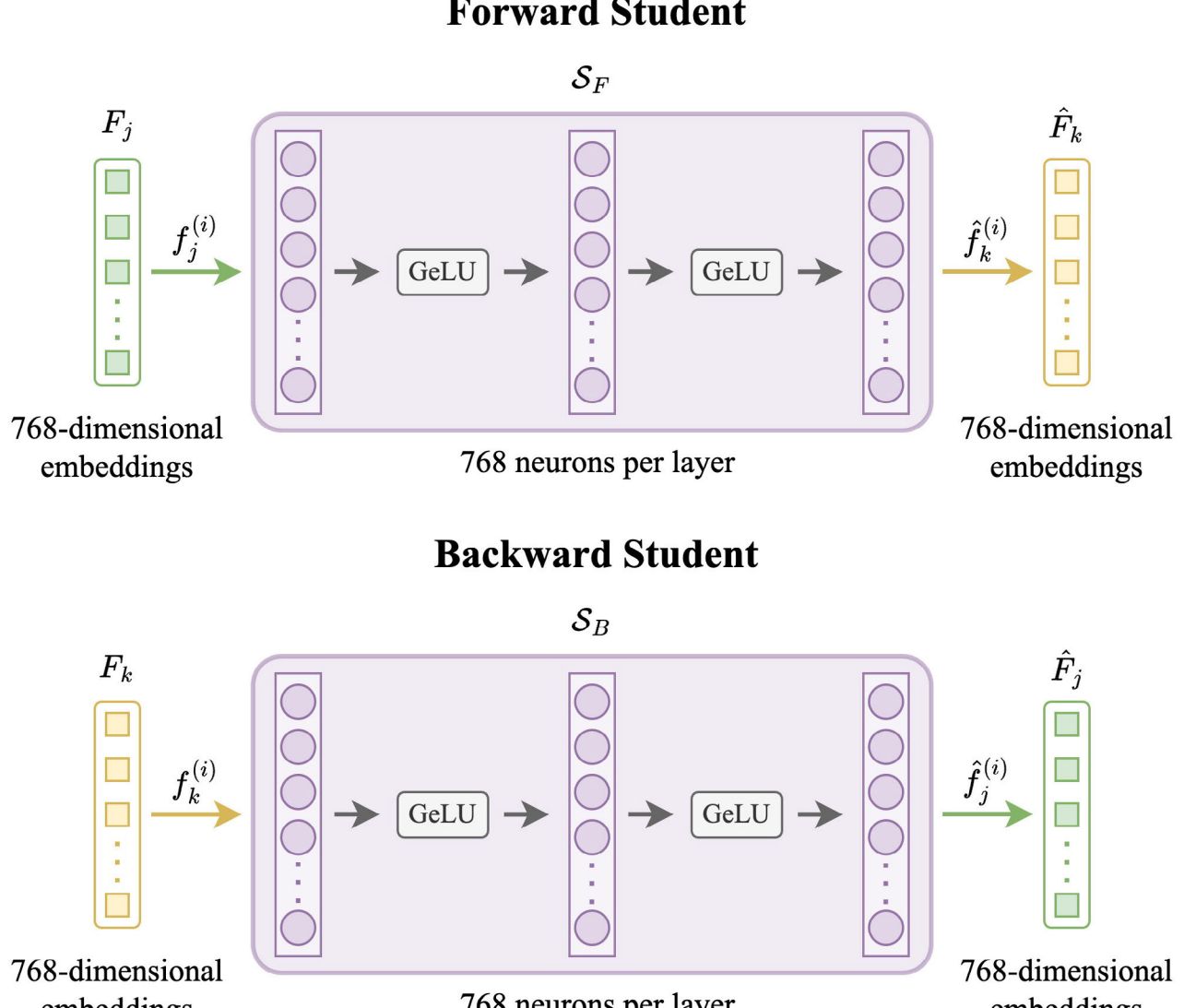

**FIGURE 5. Exploded view of the Students network architectures.**

#### 3) ATTRIBUTION OF EXISTING ASSETS

For all the competitors [1], [7], [27], except EfficientAD [8], we employed their official implementations. As far as it concerns EfficientAD, which does not provide an official repository, we leverage an implementation that obtains the most similar results with respect to the values reported in their manuscript [8]. In particular:

- PatchCore: https://github.com/amazon-science/patchcore-inspection released under Apache License 2.0;
- SimpleNet: https://github.com/DonaldRR/SimpleNet released under MIT License;

**TABLE 1. I-AUROC, AUPRO30@% and AUPRO5@% on VisA and MVTec AD for several IADS methods. Average metrics across all. Best results in bold, runner-ups underlined. All methods process high-resolution images for both training and inference. The evaluation follows the protocol described in Sec. IV-B.**

| Algorithm | VisA | | | MVTec AD | | |
|---|---|---|---|---|---|---|
| | I-AUROC | AUPRO@30% | AUPRO@5% | I-AUROC | AUPRO@30% | AUPRO@5% |
| PatchCore | **0.982** | 0.752 | 0.542 | 0.983 | 0.937 | 0.701 |
| SimpleNet | 0.904 | 0.718 | 0.469 | 0.968 | 0.914 | 0.707 |
| EfficientAD | 0.968 | 0.937 | 0.777 | 0.965 | 0.920 | 0.757 |
| RD++ | 0.930 | 0.907 | 0.758 | 0.915 | 0.901 | 0.716 |
| Ours | 0.964 | **0.952** | **0.787** | **0.988** | **0.945** | **0.782** |

- RD++: https://github.com/tientrandinh/Revisiting-Reverse-Distillation released under MIT License;
- EfficientAD: https://github.com/nelson1425/EfficientAD released under Apache License 2.0.

The VisA dataset [9] is released under the Creative Commons Attribution (CC BY 4.0) license. The MVTec AD dataset [11] is released under the Creative Commons Attribution-NonCommercial-ShareAlike 4.0 International License (CC BY-NC-SA 4.0).

## V. EXPERIMENTS

### A. ANOMALY DETECTION AND SEGMENTATION

To avoid or minimise downsampling and allow a fair evaluation via the protocol described in Sec. IV-B, at both training and inference times, we provide as input to each method the images at the highest resolution that enables execution on a single GPU. In particular, we could handle images as large as $1036 \times 1036$ pixels with EfficientAD, RD++, and our method, while the highest input resolution for PatchCore and SimpleNet was found to be $512 \times 512$ pixels. The anomaly detection and segmentation results obtained on the VisA and MVTec AD datasets are reported in Tab. 1. According to the standard metrics adopted in literature, our approach achieves the best segmentation results on VisA, with 0.952 AUPRO@30% and 0.787 AUPRO@5% and yields the state-of-the-art in both detection and segmentation on the MVTec AD dataset, with 0.988 I-AUROC, 0.945 AUPRO@30%, and 0.782 AUPRO@5%. Regarding detection performance on VisA, our method attains results comparable to the runner-up (0.968 for EfficientAD vs. 0.964 for Ours).

In fig. 6, we show some qualitative results on VisA and MVTec AD. Our method provides more accurately localised and sharper anomaly scores compared to EfficientAD [8], i.e., the second-best method in both segmentation metrics on VisA and second-best in AUPRO@5% on MVTec AD.

### B. SEGMENTATION ACROSS DEFECT SIZES AND SPEED

In Tab. 2 we report the quartile-based metrics defined in Sec. IV-A to assess segmentation performance across anomaly sizes alongside inference times, considering both high and low-resolution images. For high-resolution inputs, we rely on the same protocol as in Tab. 1, whereas, in the low-resolution setting, we adopt the official input resolution of each competitor (e.g., $224 \times 224$ or $256 \times 256$), and an

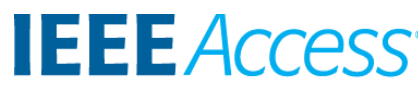

**TABLE 2.** Quartiles-based segmentation metrics and inference times at different input resolutions. Results on VisA and MVTec AD evaluating high-resolution ground truths. Inference times in ms per sample. Best results in **bold**, runner-ups underlined. IADS methods trained at **High Res** (bottom) have been fed with the highest resolution possible with our available hardware.

VisA

| | Algorithm | Input Res. | Inf. Time | AUPRO@30% | | | | | AUPRO@5% | | | | |
|---|---|---|---|---|---|---|---|---|---|---|---|---|---|
| | | | | $Q_1$ | $Q_2$ | $Q_3$ | $Q_4$ | $\rho$@30% | $Q_1$ | $Q_2$ | $Q_3$ | $Q_4$ | $\rho$@5% |
| Low Res | PatchCore | 224 × 224 | 87.151 | 0.739 | 0.741 | 0.760 | 0.779 | 0.715 | 0.443 | 0.441 | 0.471 | 0.508 | 0.405 |
| | SimpleNet | 224 × 224 | 210.833 | 0.650 | 0.654 | 0.671 | 0.690 | 0.627 | 0.309 | 0.311 | 0.338 | 0.372 | 0.275 |
| | EfficientAD | 256 × 256 | 7.837 | 0.876 | 0.904 | 0.919 | 0.931 | 0.853 | 0.646 | 0.663 | 0.697 | 0.732 | 0.603 |
| | RD++ | 256 × 256 | 17.748 | 0.770 | 0.787 | 0.814 | 0.843 | 0.733 | 0.411 | 0.429 | 0.478 | 0.541 | 0.352 |
| | Ours | 224 × 224 | **1.321** | 0.809 | 0.815 | 0.822 | 0.831 | 0.787 | 0.652 | 0.658 | 0.667 | 0.688 | 0.700 |
| High Res | PatchCore | 512 × 512 | 227.230 | 0.703 | 0.720 | 0.740 | 0.752 | 0.679 | 0.484 | 0.492 | 0.518 | 0.542 | 0.454 |
| | SimpleNet | 512 × 512 | 560.17 | 0.658 | 0.668 | 0.696 | 0.718 | 0.627 | 0.390 | 0.399 | 0.435 | 0.469 | 0.351 |
| | EfficientAD | 1036 × 1036 | 82.367 | 0.890 | 0.923 | 0.933 | 0.937 | 0.873 | 0.693 | 0.741 | 0.763 | 0.777 | 0.662 |
| | RD++ | 1036 × 1036 | 63.176 | 0.867 | 0.898 | 0.906 | 0.907 | 0.853 | 0.710 | 0.740 | 0.755 | 0.758 | 0.692 |
| | Ours | 1036 × 1036 | 1.786 | **0.935** | **0.941** | **0.946** | **0.952** | **0.926** | **0.730** | **0.749** | **0.768** | **0.787** | **0.702** |

MVTec AD

| | Algorithm | AUPRO@30% | | | | | AUPRO@5% | | | | |
|---|---|---|---|---|---|---|---|---|---|---|---|
| | | $Q_1$ | $Q_2$ | $Q_3$ | $Q_4$ | $\rho$@30% | $Q_1$ | $Q_2$ | $Q_3$ | $Q_4$ | $\rho$@5% |
| Low Res | PatchCore | 0.910 | 0.921 | 0.925 | 0.929 | 0.902 | 0.617 | 0.645 | 0.662 | 0.679 | 0.590 |
| | SimpleNet | 0.856 | 0.868 | 0.876 | 0.884 | 0.843 | 0.482 | 0.512 | 0.535 | 0.559 | 0.450 |
| | EfficientAD | 0.931 | 0.938 | 0.941 | 0.942 | 0.927 | 0.708 | 0.734 | 0.743 | 0.746 | 0.705 |
| | RD++ | 0.947 | 0.926 | 0.922 | 0.905 | 0.883 | 0.785 | 0.759 | 0.752 | 0.722 | 0.693 |
| | Ours | 0.911 | 0.912 | 0.914 | 0.914 | 0.909 | 0.633 | 0.651 | 0.660 | 0.658 | 0.625 |
| High Res | PatchCore | 0.924 | 0.932 | 0.935 | 0.937 | 0.918 | 0.653 | 0.677 | 0.691 | 0.701 | 0.633 |
| | SimpleNet | 0.913 | 0.917 | 0.917 | 0.914 | 0.914 | 0.689 | 0.705 | 0.708 | 0.707 | 0.684 |
| | EfficientAD | 0.922 | 0.925 | 0.925 | 0.920 | 0.920 | 0.758 | 0.769 | 0.767 | 0.757 | 0.760 |
| | RD++ | 0.946 | 0.922 | 0.918 | 0.901 | 0.952 | 0.782 | 0.752 | 0.744 | 0.716 | 0.684 |
| | Ours | **0.958** | **0.948** | **0.947** | **0.945** | **0.986** | **0.806** | **0.798** | **0.795** | **0.782** | **0.783** |

**TABLE 3.** Few-shot IADS performance. Best results in **bold**, runner-ups underlined.

| Algorithm | Full | | | 50-shot | | | 10-shot | | | 5-shot | | |
|---|---|---|---|---|---|---|---|---|---|---|---|---|
| | I-AUROC | AUPRO@30% | AUPRO@5% | I-AUROC | AUPRO@30% | AUPRO@5% | I-AUROC | AUPRO@30% | AUPRO@5% | I-AUROC | AUPRO@30% | AUPRO@5% |
| PatchCore | **0.982** | 0.752 | 0.542 | **0.959** | 0.724 | 0.485 | **0.948** | 0.704 | 0.459 | **0.916** | 0.698 | 0.455 |
| SimpleNet | 0.896 | 0.718 | 0.469 | 0.917 | 0.758 | 0.430 | 0.883 | 0.725 | 0.408 | 0.862 | 0.691 | 0.377 |
| EfficientAD | 0.968 | 0.937 | 0.777 | 0.831 | 0.854 | 0.569 | 0.816 | 0.806 | 0.469 | 0.810 | 0.834 | 0.511 |
| RD++ | 0.930 | 0.907 | 0.758 | 0.776 | 0.861 | 0.563 | 0.615 | 0.733 | 0.303 | 0.555 | 0.654 | 0.253 |
| Ours | 0.964 | **0.952** | **0.787** | 0.927 | **0.934** | **0.743** | 0.897 | **0.910** | **0.695** | 0.879 | **0.901** | **0.678** |

input size of 224 × 224 for our method. To measure inference times, we use the same machine for all methods and compute the average across all test samples of the two datasets. In particular, for each sample, we measure the time in ms from when it is available on the GPU to the computation of the anomaly map, after a GPU warm-up, synchronising all threads before estimating the total inference time.

We first focus on the VisA dataset, which features a much larger spread in defect sizes as well as a higher number of small defects (fig. 4). We notice how defect size impacts segmentation metrics, as, for all methods and associated input resolution, $Q_1$ is consistently inferior to $Q_4$ for both AUPRO@30% and AUPRO@5%. We then observe that our method, fed with high-resolution images, provides the best results on VisA across all quartile-based metrics while running way faster than all competitors. In particular, we highlight the significant gap w.r.t. the second-best competitor in the ability to segment accurately the smallest defects ($Q_1$-AUPRO@30%: 0.935 Ours vs. 0.890 EfficientAD, $Q_1$-AUPRO@5%: 0.730 Ours vs. 0.710 RD++) and the robustness across defect sizes ($\rho$@30%: 0.926, $\rho$@5%: 0.702). We also point out that PatchCore, which achieves better detection performance on VisA (Tab. 1, I-AUROC: 0.982 vs. 0.964), falls well behind our method in terms of segmentation performance on tiny defects ($Q_1$-AUPRO@30%: 0.739 vs. 0.935, $Q_1$-AUPRO@5%: 0.484 vs. 0.730), robustness to defect size ($\rho$@30%: 0.715 vs. 0.926, $\rho$@5%: 0.454 vs. 0.702) and inference speed (87.151 ms or 227.230 ms vs. 1.786 ms). Similar considerations concern EfficientAD: we observe detection performance comparable to our method (Tab. 1, I-AUROC: 0.968) alongside neatly inferior segmentation of small defects, robustness across sizes, and a much lower speed.

Then, still considering VisA and comparing methods across input resolutions, we note that segmentation performance tends to improve in the high-resolution setup compared to the low-resolution one, especially when considering the stricter metrics associated with AUPRO@5%. Indeed, with the exception of those related to AUPRO@30% for PatchCore, and $\rho$@30% for SimpleNet, when processing high-resolution inputs, all metrics turn out better for all methods. As shown in fig. 7, the benefits provided by the capability to process high-resolution inputs are particularly evident in the metrics aimed at assessing segmentation of the smallest defects and robustness to defect size, with methods such as RD++, EfficientAD and ours, exhibiting a performance boost thanks to the high-resolution setup. For instance, $Q_1$-AUPRO@5% increases from 0.411 to 0.710 for RD++, from 0.646 to 0.693 for EfficientAD and from 0.652 to 0.730 for our method. Similar improvements can be observed in $Q_1$-AUPRO@30% (e.g. from 0.809 to 0.935 for our method), $\rho$@5% (e.g. from 0.352 to 0.692 for RD++) and $\rho$@30% (e.g. from 0.787 to 0.926 for our method). In fig. 8, we display some anomaly maps obtained by RD++, EfficientAD and our method when fed with either low or high-resolution inputs, thereby highlighting the benefits provided by the latter set-up also from a qualitative perspective.

Moving to the results on MVTec AD reported in Tab. 2, we can notice that, compared to VisA, the metrics concerning $Q_1$ and $\rho$ tend to be quite consistently higher for all methods and input resolutions. This highlights how, due to its defect size distribution (fig. 4), MVTec AD is less representative of the challenges related to tiny anomalies and provides stable performance across sizes. Nonetheless, we can observe that, also for MVTec AD, our method fed with high-resolution inputs turns out neatly the best choice, providing the highest scores for all metrics. Yet, unlike in VisA, the runner-up competitor may not always be a method processing a high-resolution input, mainly as pertains to the less stringent metrics dealing with AUPRO@30%. Indeed, unlike in VisA, RD++ and EfficientAD exhibit a less coherent improvement trend due to the high-resolution setup.

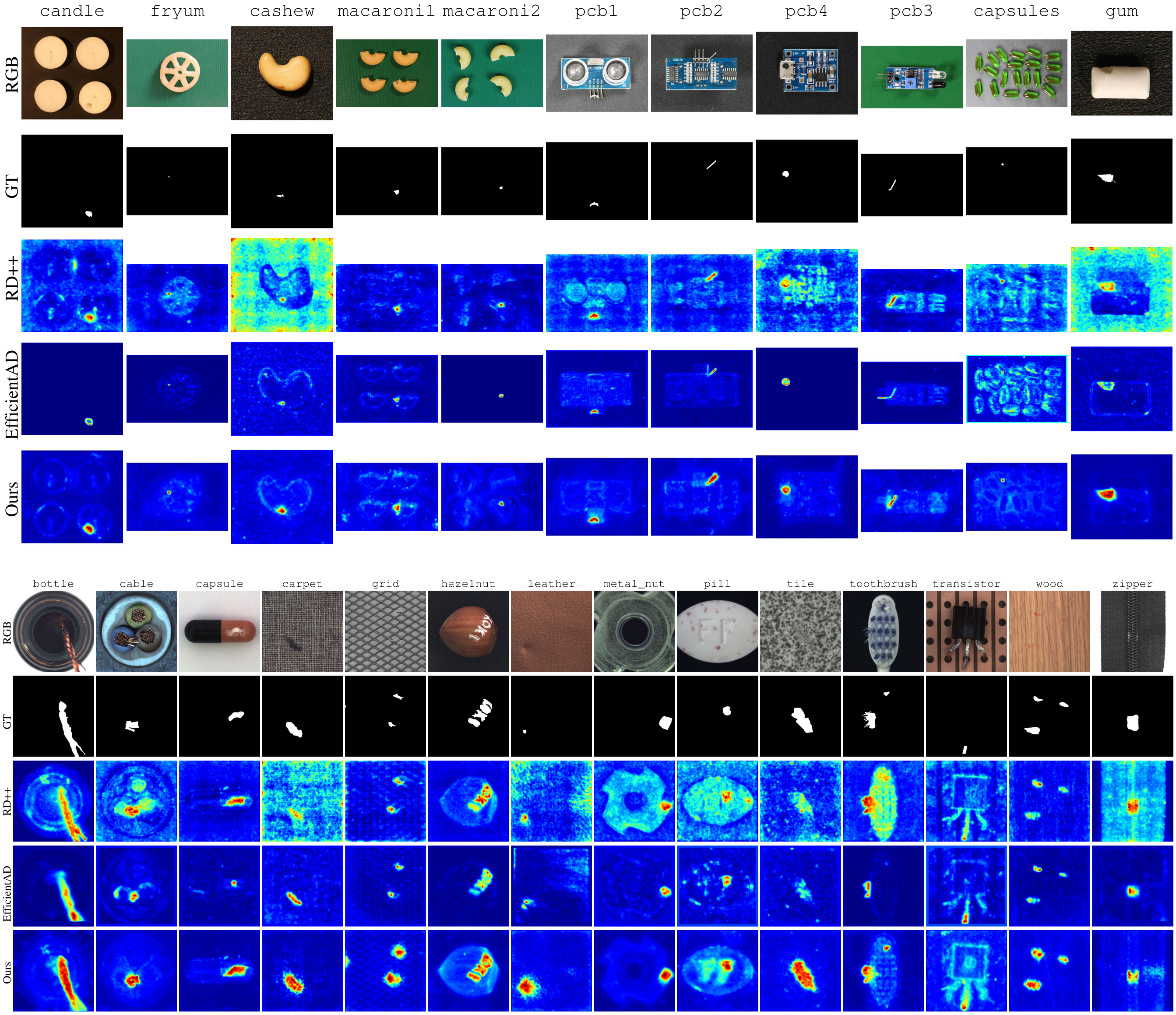

**FIGURE 6.** Qualitative results from VisA (top) and MVTec AD (bottom). All methods are trained and tested at high resolution.

### C. IMPACT OF THE INPUT RESOLUTION ON PERFORMANCE AND INFERENCE TIMES

The results in Tab. 2 emphasise the key role of the input resolution in enabling the localisation of tiny defects and the consistent segmentation across defect sizes. Indeed, our method achieves the best results on all metrics when processing high-resolution images. Moreover, our method's inference time scales gracefully with resolution. Indeed, processing high-resolution images with our method requires 1.786ms, which is only marginally superior to the time needed to process low-resolution images, i.e., 1.321ms, likely due to the small size of the Student networks. In contrast, the inference times of other networks increase significantly as resolution rises. For example, EfficientAD, the fastest competing method, has an inference time of 7.837ms on $224 \times 224$ images but requires 63.176ms — approximately 9 times longer — when processing $1036 \times 1036$ images. Finally, we emphasise that our method can process high-resolution images significantly faster than all prior methods, regardless of the input resolution.

### D. STANDARD ANOMALY SEGMENTATION PERFORMANCE

We also report in Tab. 4 the segmentation performance based on the P-AUROC metric on the VisA and MVTec AD datasets. We note that our method also achieves state-of-the-art performance on this metric in both datasets. Moreover, we wish to highlight that, unlike AUPRO@30% and AUPRO@5%, P-AUROC cannot be used to define metrics aimed at assessing the ability to localise small defects and robustness to anomaly size, such as, e.g., our novel metrics introduced in Sec. IV-A. Indeed, P-AUROC judges the per-pixel classification performance without considering to which blob, i.e. either large or small, any defective pixel in the ground truth mask belongs to. As such, a method may score a high P-AUROC by perfectly segmenting large defects while totally missing tiny ones.

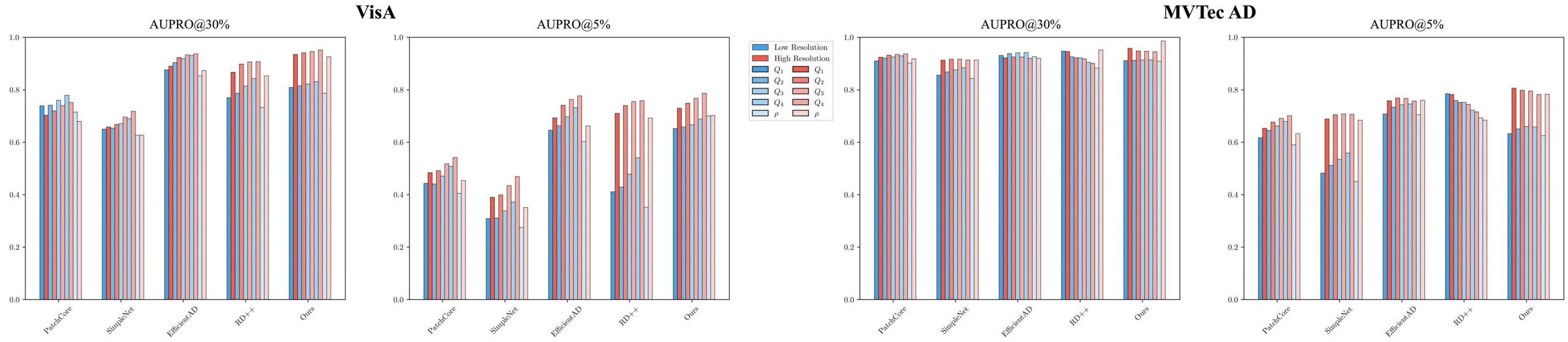

**FIGURE 7.** **Low-resolution vs. high-resolution segmentation performance. The chart highlights a general improvement in terms of segmentation performance when running the methods at higher resolution. It is evident that VisA, since it contains smaller defects, is more impacted.**

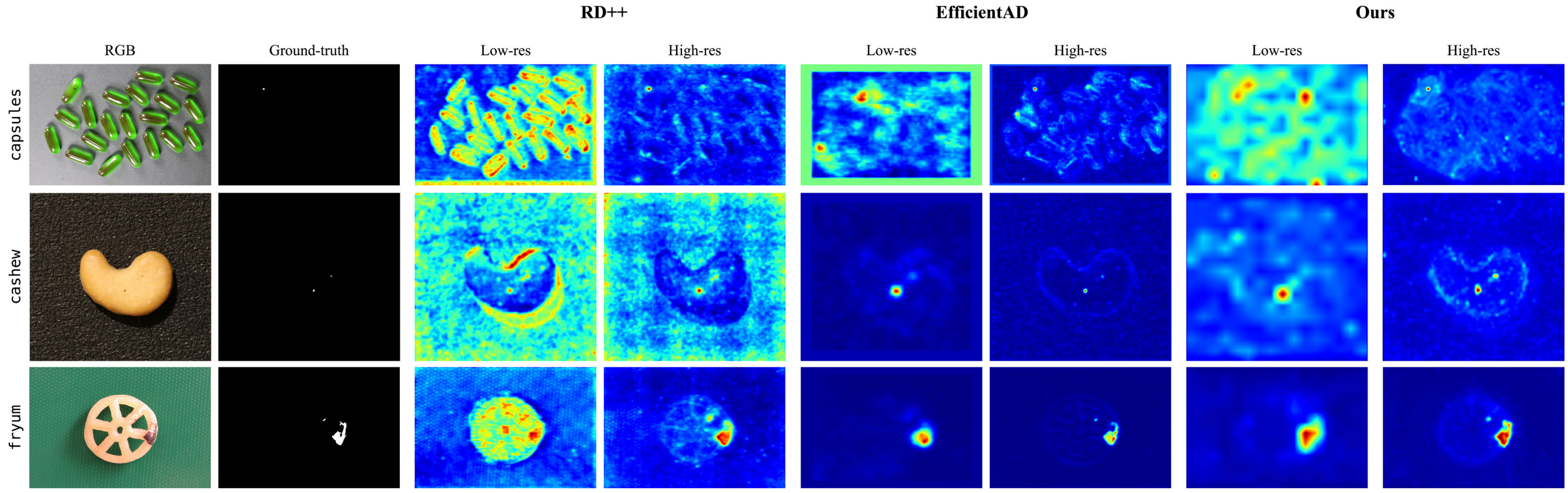

**FIGURE 8.** **Low-resolution vs. high-resolution set-up (VisA dataset). The displayed anomaly maps show that employing high-resolution inputs begets better results in detecting and segmenting small anomalies.**

**TABLE 4.** **Segmentation metrics on VisA and MVTec AD with P-AUROC metric. Best results in bold, runner-ups underlined.**

| Algorithm | VisA | MVTec AD |
|---|---|---|
| PatchCore | 0.902 | 0.980 |
| SimpleNet | 0.956 | 0.967 |
| EfficientAD | 0.977 | 0.969 |
| RD++ | 0.938 | 0.945 |
| Ours | **0.991** | **0.982** |

**TABLE 5.** **Results on VisA with different backbones. Best in bold.**

| Algorithm | Backbone | Input Resolution | AUPRO@30% | AUPRO@5% |
|---|---|---|---|---|
| Ours | ViT/B-16 | 224 × 224 | 0.868 | 0.610 |
| Ours | DINO-v2 | 224 × 224 | 0.831 | 0.688 |
| Ours | DINO-v2 | 1036 × 1036 | **0.952** | **0.787** |

### E. FEW-SHOT ANOMALY DETECTION AND SEGMENTATION

In a variety of industrial scenarios, collecting numerous nominal samples may be extremely expensive or unfeasible. Also, frequent production changeover requires fast adaptation. For these reasons, a beneficial property of a method concerns the ability to create a model of the nominal data even with few samples. Thus, we define a few-shot setting based on the VisA dataset by randomly selecting 5, 10, and 50 training samples from each category. We train our method and the competitors [1], [7], [8], [27] on these reduced training sets and then evaluate on the whole VisA test set according to the protocol proposed in Sec. IV, reporting the results in Tab. 3. Our method obtains the best segmentation performance for both metrics (AUPRO@30% and AUPRO@5%) in all the few-shot settings, significantly outperforming the competitors. For instance, with the stricter metric (AUPRO@5%) and two more challenging settings, the gap versus the runner-up is about 22% (10-shot) and 16% (5-shot). Besides, our method shows stable segmentation performance (AUPRO@30% always above 0.9) across settings. These results confirm our method's ability to optimise the nominal model even by a few images, thanks to the Student networks being shared across patches.

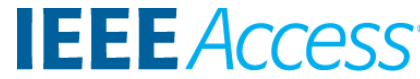

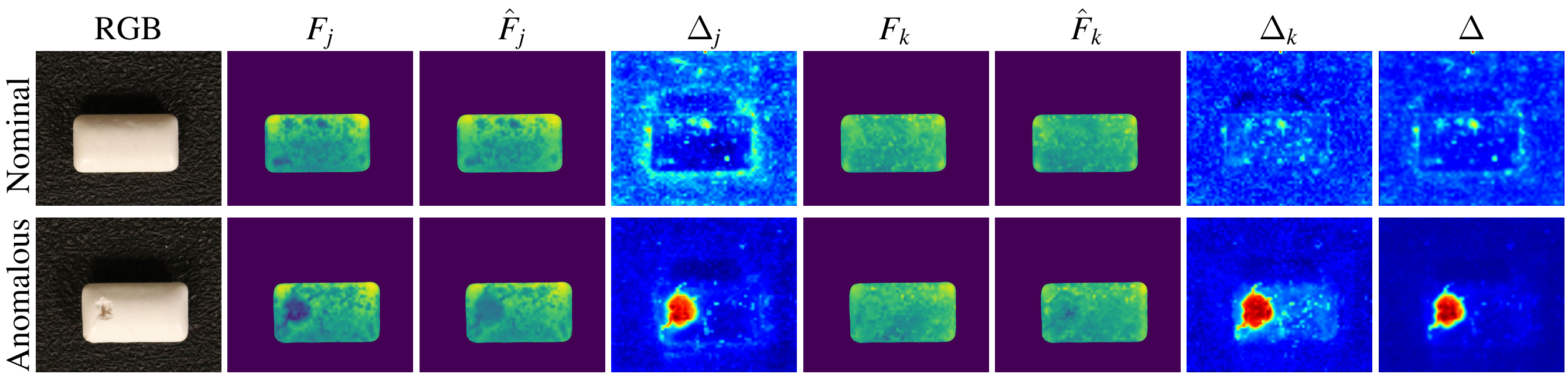

**FIGURE 9.** Features visualization. To visualize features, per-patch averages across channels are displayed by the *viridis* colormap.

### F. BACKBONE ABLATION

All previous results were achieved with our method equipped with DINO-v2 as Teacher. However, since our formulation is general, we have explored different Transformer backbones, such as ViT/B-16 pre-trained on ImageNet, and report the segmentation results on VisA in Tab. 5. As this network was pre-trained on 224 × 224 images by a classification objective, we resize images to 224 × 224 at inference time. We observe a performance drop in our method compared to processing high-resolution images (last row) with DINO-v2. Despite this, our method is still effective, outperforming all competitors except EfficientAD when operating on 224 × 224 resolution images (see Tab. 2, VisA, $Q_4$-AUPRO@30% and $Q_4$-AUPRO@5%). This demonstrates that our method can deliver competitive results with different backbones, although optimal performance is achieved by processing high-resolution images. One could, perhaps, ascribe our superior performance to the larger dataset used to pre-train DINO-v2 (∼142M images), compared to ImageNet (∼14M images), which was used to train ViT/B-16. However, if we compare the performance with DINO-v2 at 224 × 224 resolution to DINO-v2 at 1036 × 1036, we notice a significant drop in performance, even larger than the drop obtained when using ViT-B/16 in terms of AUPRO@30%. This suggests that excellent segmentation performance is primarily driven by the ability to process high-resolution images rather than the dataset size and training objective of the adopted frozen backbone.

However, it is interesting to investigate whether the effectiveness and superior performance of our method stem mainly from the adoption of DINO-v2, i.e. a strong vision foundation model, as backbone, with the contribution provided by our novel Teacher–Student paradigm turning out somehow minor. Hence, we experiment with using DINO-v2 together with PatchCore [1] and SPADE [15], i.e. two popular IADS methods which can easily accommodate changes to their backbone without requiring any ad-hoc modifications. The segmentation results on VisA, reported in Tab. 6, suggest that neither PatchCore nor SPADE can fully benefit from processing high-resolution images (1036 × 1036) using DINO-v2 as feature extractor: indeed both score lower metrics compared to the standard configuration based on a convolutional backbone (WideResNet101) fed with low-resolution inputs (224 × 224) and, given the same DINO-v2 backbone, perform much worse than our method (last row in Tab. 6). Hence, the key to the effectiveness of our method is the combination of DINO-v2 features with the proposed Teacher–Student paradigm, as the former alone cannot be deployed as effectively via the memory bank mechanism adopted in PatchCore and SPADE. Eventually, we further validate the importance of our Teacher–Student paradigm by learning the Students from a Transformer backbone geared towards low-resolution images, namely ViT/B-16 trained on ImageNet (the same as in Tab. 5 of the main paper): our Teacher–Student approach realized with a backbone different from DINO-v2 neatly outperforms Patchcore and SPADE, both in their original formulation or when equipped with DINO-v2.

**TABLE 6.** Segmentation metrics on VisA. Best results in bold, runner-ups underlined.

| Algorithm | Backbone | Input Resolution | AUPRO@30% | AUPRO@5% |
|---|---|---|---|---|
| PatchCore | WideResNet101 | 224 × 224 | 0.779 | 0.508 |
| PatchCore | DINO-v2 | 1036 × 1036 | 0.705 | 0.445 |
| SPADE | WideResNet101 | 224 × 224 | 0.780 | 0.480 |
| SPADE | DINO-v2 | 1036 × 1036 | 0.779 | 0.462 |
| Ours | ViT/B-16 | 224 × 224 | <u>0.868</u> | <u>0.610</u> |
| Ours | DINO-v2 | 1036 × 1036 | **0.952** | **0.787** |

### G. FEATURES VISUALISATION

In fig. 9, we consider a nominal (top) and anomalous (bottom) test image from VisA and show the feature computed by the Teacher, $F_j, F_k$, those predicted by the Students, $\hat{F}_j, \hat{F}_k$, as well as the corresponding difference $\Delta_j, \Delta_k$ and final anomaly maps, $\Delta$. We can notice how, for the nominal image, the predicted features look similar to the actual ones, which results in low anomaly scores. For the anomalous image, instead, the Students fail to predict the actual features computed by the Teacher due to these containing embeddings that fall out of their training distribution. Thus, the distances between the actual and predicted features enable pinpointing anomalies in both $\Delta_j, \Delta_k$, with the final anomaly map computed by product-based aggregation exhibiting a more accurate localisation of the anomaly score peak and less noise.

### H. ABLATION ON THE LAYERS CONSIDERED FOR THE FORWARD AND BACKWARD FEATURE TRANSFER NETWORKS

We investigate the impact of the pair of layers, i.e., $j$ and $k$ (Sec. III), between which the Students learn to predict the feature computed by the Teacher. In Tab. 7, we report results for various choices of layer pairs. The notation $[j, k]$ signifies that the anomaly maps associated with the chosen layers are fused, i.e., as described in the main paper, $\Delta = \Delta_j \cdot \Delta_k$. With $[j \rightarrow k]$ and $[k \leftarrow j]$, we intend, instead, the performance yielded by the individual anomaly maps, namely $\Delta_k$ and $\Delta_j$, respectively. By comparing the results provided by the first three pairs listed in Tab. 7, we can note that mapping features between deeper layers begets better detection and segmentation results, with the last four layers ($j = 8, k = 12$) providing the best performance. Then, analysing also the last two choices considered in the Table, we observe that learning to map features from closer layers, such as $j = 11$ and $k = 12$, yields a significant performance drop. Interestingly, a recent paper has provided evidence that the function learned by a single Transformer layer is smooth [31]. This leads us to conjecture that the tasks of predicting the embeddings back and forth between two adjacent layers may be simpler than between layers spaced further apart, with the Students having more chances to learn functions that generalise to anomalous samples. We also point out that, beyond layer 8, when considering a sufficiently large gap between the layers in the pair, performance is relatively stable. Finally, we notice that fusing the maps obtained from the forward and backward mappings tends to provide better performance quite consistently, except for layers $[1, 4]$. We conjecture that it is possible to select the best layers for the forward and backward feature transfer networks by selecting the ones that yield the lowest reconstruction error at training time.

**TABLE 7.** Ablation on the pair of layers. Best results in bold, runner-ups underlined.

| Layers | I-AUROC | AUPRO@30% | AUPRO@5% |
|---|---|---|---|
| $[1, 4]$ | 0.906 | 0.828 | 0.570 |
| $[1 \rightarrow 4]$ | 0.913 | 0.906 | 0.682 |
| $[1 \leftarrow 4]$ | 0.773 | 0.663 | 0.378 |
| $[4, 8]$ | 0.940 | 0.941 | 0.773 |
| $[4 \rightarrow 8]$ | 0.924 | 0.942 | 0.764 |
| $[4 \leftarrow 8]$ | 0.931 | 0.903 | 0.702 |
| $[8, 12]$ | **0.964** | **0.952** | **0.787** |
| $[8 \rightarrow 12]$ | 0.953 | 0.943 | 0.773 |
| $[8 \leftarrow 12]$ | 0.949 | 0.925 | 0.745 |
| $[10, 12]$ | 0.960 | 0.950 | 0.784 |
| $[10 \rightarrow 12]$ | 0.957 | 0.926 | 0.742 |
| $[10 \leftarrow 12]$ | 0.960 | 0.947 | 0.782 |
| $[11, 12]$ | 0.956 | 0.946 | 0.774 |
| $[11 \rightarrow 12]$ | 0.868 | 0.853 | 0.710 |
| $[12 \leftarrow 11]$ | 0.888 | 0.876 | 0.730 |

### I. ABLATION ON THE DISTANCES USED AT TRAINING AND INFERENCE TIMES

Tab. 8 reports the results obtained by considering different choices for the distance function used to optimise the Students, that is, the per-patch loss function, as well as to compute the two anomaly maps associated with the forward and backward predictions. In particular, for both training and inference, we consider the cosine and $\ell_2$ distances and report the results of the four possible combinations. Our choice (first row) shows slightly better performance, though the differences are very small.

**TABLE 8.** Ablation on distances used at training and inference times. Best results in bold, runner-ups underlined.

| Training | Inference | I-AUROC | AUPRO@30% | AUPRO@5% |
|---|---|---|---|---|
| Cosine distance | $\ell_2$ distance | **0.964** | **0.952** | 0.787 |
| $\ell_2$ distance | $\ell_2$ distance | 0.954 | 0.950 | 0.786 |
| Cosine distance | Cosine distance | 0.957 | **0.952** | **0.790** |
| $\ell_2$ distance | Cosine distance | 0.954 | 0.938 | 0.741 |

### J. ABLATION ON THE FUNCTION EMPLOYED TO FUSE THE ANOMALY MAPS

Using the best combination of layers ($j = 8$ and $k = 12$, as shown in Tab. 7, we investigate on the fusion strategy. In particular, we adopt the product with the goal of minimizing potential false positives from the maps produced by each of the two Students. Indeed, the product can be viewed as a logical AND between the two maps, so that a pixel is highlighted as anomalous only if both Students agree to predict so. As shown in Tab. 9, choosing the product as the aggregation function enhances the performance of the individual maps, while addition slightly degrades their performance.

**TABLE 9.** Ablation on aggregation of anomaly maps. $j = 8, k = 12$. Best results in bold, runner-ups underlined.

| Anomaly map | I-AUROC | AUPRO@30% | AUPRO@5% |
|---|---|---|---|
| $\Delta_k \cdot \Delta_j$ | **0.964** | **0.952** | **0.787** |
| $\Delta_k + \Delta_j$ | 0.944 | 0.931 | 0.732 |
| $\Delta_j$ | 0.953 | 0.943 | 0.773 |
| $\Delta_k$ | 0.949 | 0.925 | 0.745 |

### K. TRAINING TIME

We provide in Tab. 10 the average time in hours needed per class to train every framework, given the number of epochs reported in their official implementations. These timings have been computed using the same hardware employed for all our experiments.

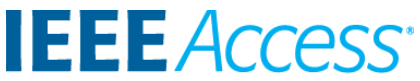

**TABLE 10. Training time required on the VisA dataset. Average training time in hours per class. All methods are trained at high resolution.**

| Algorithm | PatchCore | SimpleNet | EfficientAD | RD++ | Ours |
|---|---|---|---|---|---|
| Training time | 1.212 | 6.266 | 7.783 | 28.767 | 2.361 |

### L. SENSITIVITY TO BACKGROUND

We show in fig. 10 a typical failure case of the proposed method. In the presence of speckles or extraneous objects in the background, these will exert a high response in the anomaly map. This is due to the fact that this noise in the background is unseen in the training data.

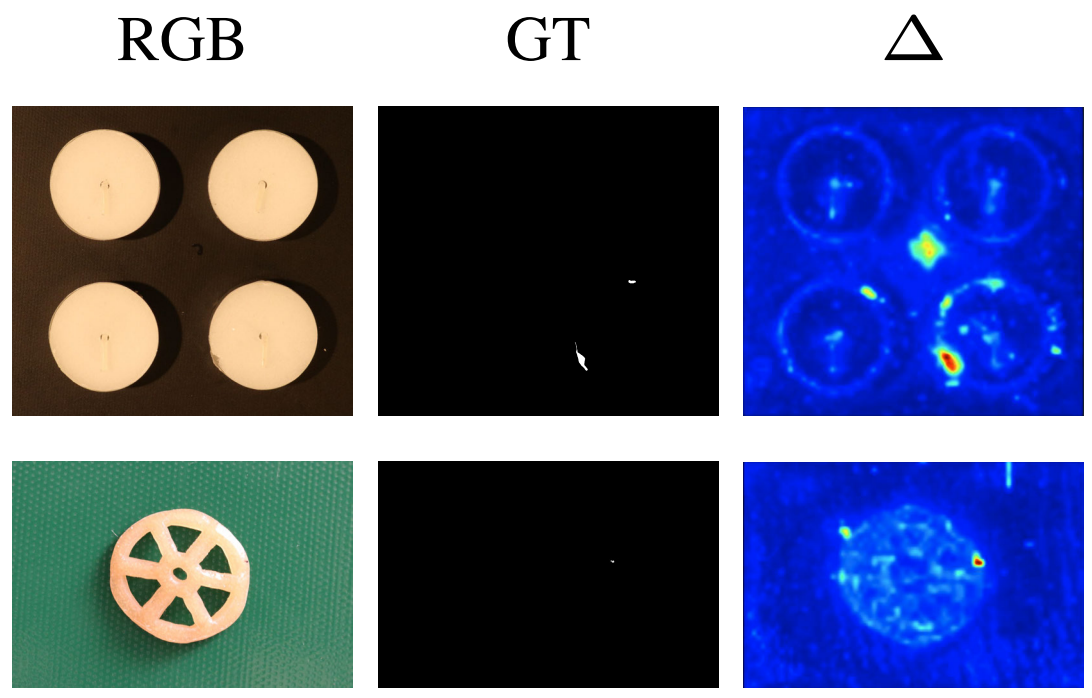

**FIGURE 10. Background noise. In the presence of background artefacts unseen at training time, these will be detected as anomalies during the inference.**

## VI. CONCLUDING REMARKS

Our novel Teacher-Student paradigm allows for processing images as large as ∼1 Mpx effectively and efficiently. As a result, our method can localise defects across the whole range of sizes featured by the main IADS benchmarks more accurately, more robustly and more rapidly than previous proposals. Besides, our paradigm is particularly amenable to challenging few-shot use cases.

However, IADS scenarios requiring to identification of minuscule flaws inspecting large surfaces, or both, may call for scrutinising way larger input images, such as, e.g. ∼10-12 Mpx, or even more. Thus, we reckon that current state-of-the-art approaches leveraging successful, possibly pre-trained, deep learning architectures may fall short in addressing the challenges posed by very high-resolution IADS. In this respect, a limitation of our method resides in the small spatial size of the output anomaly map, which is constrained by the leveraged backbone. A possible avenue for enhancement may consist of incorporating strategies to compute much denser feature maps (e.g., FeatUp [32]).

To conclude, we highlight that our experiments, together with the proposed evaluation protocol, contribute to shedding light on the significance of the input resolution in IADS. We hope that our work may pave the way for further investigation along this path, including newer benchmarks tailored towards very high-resolution use cases.

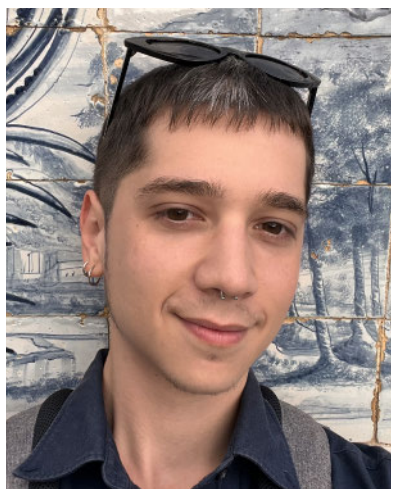

**ALEX COSTANZINO** received the bachelor's degree in automation engineering from the University of Bologna, in 2020 and the master's degree in artificial intelligence from the University of Bologna, in 2022. He is currently pursuing the Ph.D. degree in computer science and engineering with the Computer Vision Laboratory (CVLab), University of Bologna. His research interests include artificial intelligence and deep learning techniques for computer vision, in particular for depth estimation and anomaly detection, and segmentation.

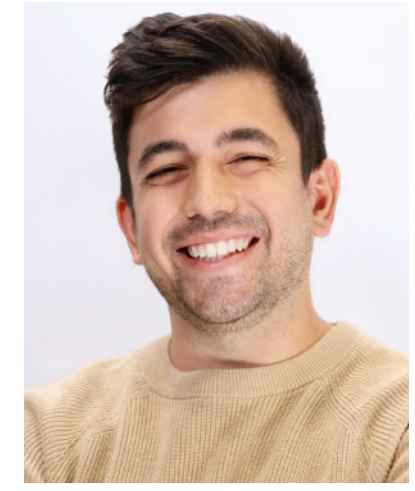

**PIERLUIGI ZAMA RAMIREZ** received the Ph.D. degree in computer science and engineering, in 2021. He is currently an Assistant Professor with the University of Bologna. He has co-authored more than 30 publications on computer vision research topics, such as semantic segmentation, depth estimation, optical flow, domain adaptation, virtual reality, and 3D computer vision.

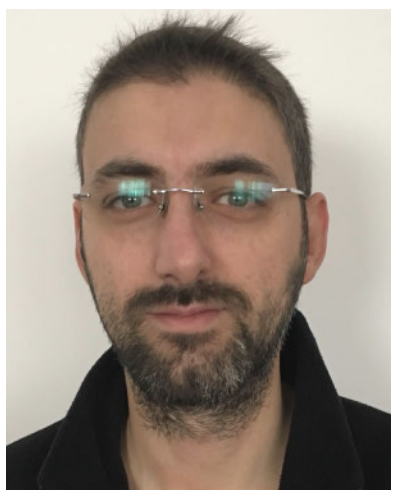

**GIUSEPPE LISANTI** is currently an Associate Professor with the Department of Computer Science and Engineering, University of Bologna. He has co-authored over 50 publications. He actively collaborates with other research centers and has participated in various roles in multiple research projects. His primary research interests include computer vision and the application of deep learning to computer vision problems. In 2017, he received the Best Paper Award from the IEEE Computer Society Workshop on Biometrics.

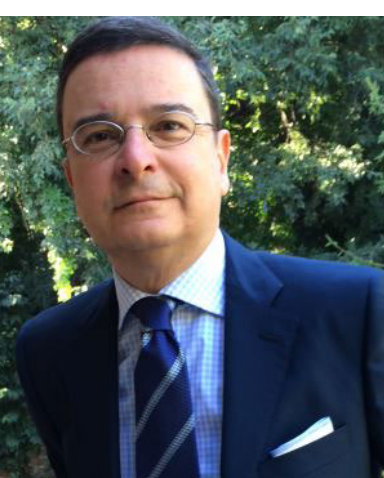

**LUIGI DI STEFANO** (Member, IEEE) received the Ph.D. degree in electronic engineering and computer science from the University of Bologna, in 1994. He was a Scientific Consultant for major companies, in the fields of computer vision and machine learning. He is currently a Full Professor with the Department of Computer Science and Engineering, University of Bologna, where he founded and leads the Computer Vision Laboratory (CVLab). He is the author of more than 150 papers and several patents. His research interests include image processing, computer vision, and machine/deep learning. He is a member of the IEEE Computer Society and the IAPR-IC.